\documentclass{esannV2}
\usepackage{multirow}
\usepackage{hyperref}
\usepackage{caption} 
\usepackage{amssymb,amsmath,array}
\usepackage{graphicx}
\begin{document}
\title{
Lightweight Cross-Modal Representation Learning}
\author{Bilal FAYE$^1$, Hanane AZZAG$^1$, Mustapha LEBBAH$^2$, Djamel BOUCHAFFRA$^3$
%
%
\vspace{.3cm}\\
%
1- LIPN, UMR CNRS 7030
Sorbonne Paris Nord University, Villetaneuse, France
%
\vspace{.1cm}\\
2- DAVID Lab, University of Versailles,
University Paris-Saclay, Versailles, France
\vspace{.1cm}\\
3- Center for Development of Advanced Technologies, Algiers, Algeria
}

\maketitle

\begin{abstract}
Low-cost cross-modal representation learning is crucial for deriving semantic representations across diverse modalities such as text, audio, images, and video. Traditional approaches typically depend on large specialized models trained from scratch, requiring extensive datasets and resulting in high resource and time costs. To overcome these challenges, we introduce a novel approach named Lightweight Cross-Modal Representation Learning (LightCRL). This method uses a single neural network titled Deep Fusion Encoder (DFE), which projects data from multiple modalities into a shared latent representation space. This reduces the overall parameter count while still delivering robust performance comparable to more complex systems. The code is available via \href{https://github.com/b-faye/lightweightCRL/}{https://github.com/b-faye/lightweightCRL/}

\end{abstract}

\section{Introduction}
Modalities are channels for exchanging information between humans and the environment, including text, audio, images, and video. Cross-modal learning bridges these channels using alignment and fusion strategies. According to~\cite{orton2020vision}, cross-modal fusion combines data into a unified representation—early, late, intermediate, or hybrid fusion. Early fusion integrates raw features at the input level, late fusion at the output level, and intermediate fusion at multiple stages. Hybrid fusion employs a mix of these methods. Cross-modal alignment aligns modalities within a shared semantic space using techniques like contrastive learning and encoder-decoder architectures ~\cite{chen2020simple}.
%
Contrastive learning enhances alignment between similar samples and reduces it for dissimilar ones, as used in ConVIRT and CLIP~\cite{radford2021learning,zhang2022convirt}. Masked modeling, like in VisualBERT, learns by predicting unseen parts of data~\cite{devlin2019bert}. Encoder-decoder structures, seen in visualGPT, facilitate modality interactions~\cite{li2021visualgpt}. These methods, however, require large models for each modality and extensive aligned datasets, posing high computational costs and data availability challenges.
\newline
\indent To address the computational and data challenges in cross-modal learning, we introduce Lightweight Cross-Modal Representation Learning (LightCRL), a method designed to acquire high-level semantic representations while minimizing computational resources efficiently. LightCRL significantly reduces the number of parameters required and operates independently of large aligned datasets by leveraging large pre-trained models that remain frozen during the process. 
At the core of LightCRL is the Deep Fusion Encoder (DFE), which stands as the unique trainable element shared across different modalities. The DFE integrates a learnable context vector specific to each modality, combining these vectors with the embeddings of the modalities to concentrate training efforts solely on the parameters of the DFE.
Our contributions are threefold: (i) we maximize efficiency by using frozen pre-trained models to curtail extensive training and conserve resources; (ii) we streamline the learning process and enhance information integration across modalities with the unified DFE, even in scenarios with limited datasets; (iii) and we enable effective differentiation and representation of diverse data through context-aware fusion within the DFE, using consistent parameters across different modalities.
%
%
\section{Proposed Method}
Lightweight Cross-Modal Representation Learning (LightCRL) offers a novel approach to efficient representation learning across modalities. It uses large pretrained encoders for each modality, kept fixed during training. LightCRL introduces a unique neural network, "Deep Fusion Encoder", to handle cross-modal representations. 
\begin{figure*}[t]
      \centering
      \includegraphics[width=1\textwidth]{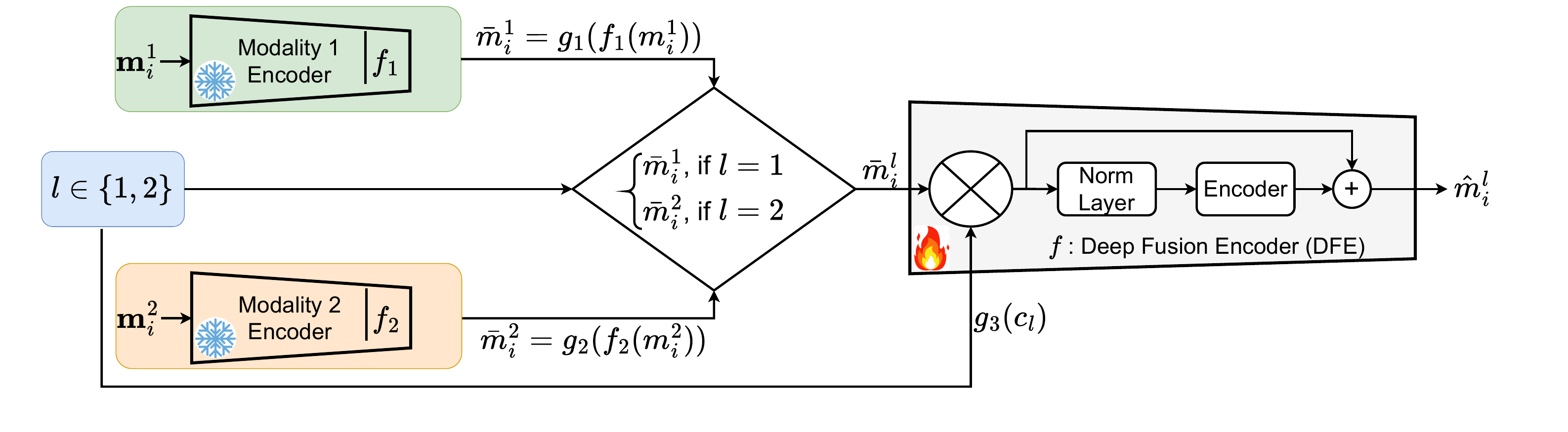}
      \caption{LightCRL framework: Only DFE denoted $f$ is trained  for cost-effectiveness, while $f_1$ and $f_2$ remain static. $\mathbf{m}^1_i$, $\mathbf{m}^2_i$ represent modalities $1$ and $2$ with their respective embedding $\mathbf{\Bar{m}}^1_i$ and $\mathbf{\Bar{m}}^2_i$, using  respective frozen encoders. $\mathbf{\hat{m}}^1_i$ and $\mathbf{\hat{m}}^2_i$ are embeddings on the common latent space. $c_l$ is the context identifier for modality $l$, and $g_1$, $g_2$, and $g_3$ ensure uniform dimensions.}
      \label{fig:fusion}
\end{figure*}
The DFE ($f$) in Figure~\ref{fig:fusion} plays a pivotal role in our framework. Nonlinear projection functions $g_1$, $g_2$, and $g_3$  ensure input dimension consistency for $f$ \cite{zhang2022convirt}. $f$ combines trainable context vector $c$ with frozen pre-trained encoders $f_1$ and $f_2$, tailored to each modality for coherent fusion. This approach allows DFE to represent diverse modalities effectively with shared parameters. We propose a shallow neural network for DFE, maintaining semantic representation efficiently and cost-effectively.
Consider datasets $\mathbf{M}_1$ and $\mathbf{M}_2$ for modalities $1$ and $2$ respectively, with data pairs $(\mathbf{m}^1_i, \mathbf{m}^2_i)$ where $\mathbf{m}^1_i$ is from modality $1$ and $\mathbf{m}^2_i$ from modality $2$. The paired dataset is $\{(\mathbf{m}^1_1, \mathbf{m}^2_1), (\mathbf{m}^1_2, \mathbf{m}^2_2), ..., (\mathbf{m}^1_N, \mathbf{m}^2_N)\}$ with $N$ samples. LightCRL, as depicted in Figure~\ref{fig:fusion}, trains $f$ for cross-modal representation via alignment in a shared latent space. 
The context modality identifier $c_l$ with $l \in \{1, 2\}$ enhances the Deep Fusion Encoder (DFE) representation. This approach enables uniform parameter utilization across different modalities, thereby eliminating the necessity to learn individual encoder weights for each modality. This method is called "Lightweight" due to its reduced computational overhead and increased efficiency in managing multiple modalities within a single model framework.
%
For each input pair $(\mathbf{m}^1_i, \mathbf{m}^2_j)$, LightCRL's transformation is defined as $\mathbf{\hat{m}}^1_i = f(g_1(f_1(\mathbf{m}^1_i)) \otimes g_3(c_1))$, and $\mathbf{\hat{m}}^2_i = f(g_2(f_2(\mathbf{m}^2_i)) \otimes g_3(c_2))$.
%
The symbol $\otimes$ represents the fusion operation between the context modality vector and embeddings from frozen encoders. Fusion strategies include element-wise methods (addition, multiplication, concatenation) outlined in~\cite{smith2020multimodal}, or cross-attention fusion~\cite{johnson2019cross}.
During training, we sample a minibatch of $K$ input pairs $(\mathbf{m}^1_i, \mathbf{m}^2_i)$ from the training dataset of $N$ pairs. 
LightCRL's training objective function includes a contrastive loss between modality $1$ and $2$ for pairs $(\mathbf{\hat{m}}^1_i, \mathbf{\hat{m}}^2_j)$:
\begin{equation}
\label{eqn:lij}
    \ell_{ij} = - \log \left(\frac{\exp(\langle \mathbf{\hat{m}}^1_i, \mathbf{\hat{m}}^2_j \rangle / \tau)}{\sum_{k=1}^K \exp(\langle \mathbf{\hat{m}}^1_i, \mathbf{\hat{m}}^2_k \rangle/\tau)}\right)
\end{equation}
\newline
\indent  The term $\langle \mathbf{\hat{m}}^1_i, \mathbf{\hat{m}}^2_i \rangle$ represents cosine similarity, with $\tau \in \mathbb{R}^+$ as a learnable temperature parameter. This loss function preserves mutual information between true pairs through representation functions. To ensure symmetry, we introduce a similar contrastive loss from modality $2$ to modality $1$: $\ell_{ji}$.
The matching pairs are situated along the diagonal of the similarity matrix $(\mathbf{\hat{m}}^1_i, \mathbf{\hat{m}}^2_i)$, which serves as the target for the loss function:
\begin{equation}
\label{eqn:tij}
    t_{ij} =  \frac{\exp((\langle \mathbf{\hat{m}}^1_i, \mathbf{\hat{m}}^1_j \rangle + \langle \mathbf{\hat{m}}^2_i, \mathbf{\hat{m}}^2_j \rangle)  / 2\cdot \tau)}{\sum_{k=1}^K \exp((\langle \mathbf{\hat{m}}^1_i, \mathbf{\hat{m}}^1_k \rangle + \langle \mathbf{\hat{m}}^2_i, \mathbf{\hat{m}}^2_k \rangle) /2 \cdot \tau)}
\end{equation}
\newline

The ultimate training loss $\mathcal{L}$ \eqref{eqn:cost} is computed by combining the two losses $\ell_{ij}$ and $\ell_{ji}$ and averaging them over all pairs within each minibatch.
\begin{equation}
\label{eqn:cost}
    \mathcal{L} = \frac{1}{2\cdot K}\sum_{i=1}^K \sum_{j=1}^K t_{ij}\cdot \ell_{ij} + t_{ji}\cdot \ell_{ji} 
\end{equation}
In LightCRL, the context modality identifier $c_l$ offers two main benefits. Firstly, it allows a unified encoder (DFE) $f$ to process embeddings from different pre-trained encoders ($f_1$ and $f_2$), enhancing semantic representation while reducing parameters. Secondly, the context modality vector $g_3(c_l)$ acts as an implicit prior, enriching cross-modal representation~\cite{wang2021you}. LightCRL's contrastive learning ensures efficient alignment across modalities, facilitating easy transfer to various tasks, thus enhancing its applicability.
\section{Experiments}
In this section, we apply LightCRL to the COCO Captions dataset~\cite{lin2014microsoft}, consisting of $330K$ images ($\textbf{M}_1$) with $5$ captions ($\textbf{M}_2$) per image.
\newline
\indent \textbf{Parameter size gain:}
In our Lightweight Cross-Modal Representation Learning (LightCRL) framework, we use pre-trained BERT (12 layers, 110M parameters) and Vision Transformer (ViT) (12 layers, 86M parameters) to encode captions, keeping these models' parameters frozen to minimize the number of trainable parameters~\cite{Devlin2018, Dosovitskiy2020}. Our training focuses exclusively on the Deep Fusion Encoder (DFE), a Transformer block with 4-head attention with just 1M parameters, employing two fusion methods: addition (DFE-add) and scaled dot-product attention (DFE-dot). This approach drastically reduces the computational burden compared to training all model parameters. The models are trained for 500 epochs with early stopping criteria.
%
%
\newline
\indent \textbf{Zero-shot image classification:} Utilizes existing capabilities to classify images without prior training on specific classes. It involves computing feature embeddings for both images and text names of all classes within the CIFAR-10 dataset. Then, the cosine similarity of these embeddings is calculated and normalized into a probability distribution using softmax.
\begin{table}[t]
    \centering
    \begin{tabular}{lccccc}
        \hline
        \multirow{3}{*}{\textbf{Model}} & \multicolumn{5}{c}{\textbf{Accuracy}} \\
        \cline{2-6}
         & top-1 & top-2 & top-3 & top-4 & top-5  \\
        \hline
        conVIRT & 62.12 & 76.53 & 84.26 & 89.22 & 92.95 \\
        DFE-add (LightCRL) & \textbf{78.26} & \textbf{88.69} & 92.5 & 94.87 & 96.48 \\
        DFE-dot (LightCRL) & 75.62 & 87.37 & \textbf{92.62} & \textbf{95.26} & \textbf{97.00} \\
        \hline
    \end{tabular}
    \caption{
    Assessment of zero-shot classification capabilities on the CIFAR-10 dataset. Here, DFE-add signifies the DFE using addition as the fusion method, and DFE-dot indicates the DFE employing scaled dot-product attention for the fusion process.
    }
    \label{tab:zero_shot}
\end{table}
Table~\ref{tab:zero_shot} demonstrates the superior performance of LightCRL models (DFE-add and DFE-dot) on CIFAR-10 without retraining. These models, pretrained on COCO Captions and adapted to CIFAR-10, outperform the resource-intensive conVIRT approach, despite CIFAR-10 containing categories absent in COCO Captions.\newline
\indent \textbf{Linear Classification}: Using pre-trained ConVIRT and DFEs models, we apply a linear classifier to encode CIFAR-100 images. Only the linear classifier is trained for $100$ epochs, while the pre-trained models remain frozen. This approach assesses the quality of extracted image features with the pre-trained DFEs. Validation on the test set occurs every $20$ epochs during training.
\begin{table}[t]
    \centering
    \begin{tabular}{lccccc}
        \hline
        \multirow{3}{*}{\textbf{Model}} & \multicolumn{5}{c}{\textbf{Epoch}} \\
        \cline{2-6}
         & 20 & 40 & 60 & 80 & 100   \\
        \hline
        conVIRT & 55.67 & 59.22 & 61.09 & 62.14 & 62.71 \\
        DFE-add (LightCRL) & \textbf{64.63} & \textbf{66.78} & \textbf{67.93} & \textbf{68.88} & \textbf{69.37} \\
        DFE-dot (LightCRL) & 62.75 & 63.37 & 64.57 & 64.98 & 65.29 \\
        \hline
    \end{tabular}
    \caption{
    Evaluating the precision of linear classification tasks on the CIFAR-100 dataset. Here, DFE-add signifies the DFE using addition as the fusion method, and DFE-dot indicates the DFE employing scaled dot-product attention for the fusion process.}
    \label{tab:linear_classification}
\end{table}
Table~\ref{tab:linear_classification} displays the superior performance of LightCRL models (DFE-add and DFE-dot) for linear classification, echoing the findings from zero-shot classification. By utilizing pre-trained DFEs without image augmentation and keeping them frozen, LightCRL models outperform the conVIRT approach. This suggests that our method yields higher-quality image features, enhancing discrimination across categories and improving classification performance.
\newline
\indent \textbf{Fine-tuning}: We follow a similar strategy as linear classification, but unlike in linear classification, we unfreeze the pre-trained ConVIRT and DFEs. This approach closely simulates real-world scenarios where pre-trained models weights are adjusted. We evaluate this technique on the Tiny ImageNet dataset, training models for $100$ epochs without using data augmentation. Validation on the test set is conducted every $20$ epochs during training.
\begin{table}[t]
    \centering
    \begin{tabular}{lccccc}
        \hline
        \multirow{3}{*}{\textbf{Model}} & \multicolumn{5}{c}{\textbf{Epoch}} \\
        \cline{2-6}
         & 20 & 40 & 60 & 80 & 100   \\
        \hline
        conVIRT & 59.00 & 62.71 & 64.45 & 65.37 & 65.89 \\
        DFE-add (LightCRL) & \textbf{64.49} & \textbf{66.31} & \textbf{67.60} & \textbf{68.12} & \textbf{68.91} \\
        DFE-dot (LightCRL) & 64.21 & 66.09 & 66.70 & 67.74 & 68.18 \\
        \hline
    \end{tabular}
    \caption{
    Assessing the accuracy of fine-tuning outcomes on the Tiny ImageNet dataset. DFE-add denotes the DFE model that employs addition as its fusion technique, while DFE-dot signifies the DFE model using scaled dot-product attention for its fusion method.}
    \label{tab:fine_tuning}
\end{table}
Table~\ref{tab:fine_tuning} confirms the results from zero-shot and linear classification experiments. 
Models pre-trained with LightCRL method offer strong feature representation, serving as robust backbones across tasks.
\section{Conclusion}
We introduce LightCRL, a cost-effective method for multimodal representation learning. LightCRL's efficiency reduces reliance on extensive datasets and long training times. It utilizes frozen pre-trained models for encoding modalities, with training focused solely on the DFE. DFE employs uniform parameters for different modalities and incorporates a context modality identifier for relevant representation. LightCRL offers a versatile and robust cross-modal representation framework, demonstrated through experiments aligning text and image modalities.
\bibliographystyle{plain}
\bibliography{bibliography} 
\end{document}